%% file: MAIN.tex
\documentclass{article}

\usepackage{PRIMEarxiv}

\usepackage[utf8]{inputenc} % allow utf-8 input
\usepackage[T1]{fontenc}    % use 8-bit T1 fonts
\usepackage{hyperref}       % hyperlinks
\usepackage{amsmath}
\usepackage{amssymb}
\usepackage{url}            % simple URL typesetting
\usepackage{booktabs}       % professional-quality tables
\usepackage{amsfonts}       % blackboard math symbols
\usepackage{nicefrac}       % compact symbols for 1/2, etc.
\usepackage{microtype}      % microtypography
\usepackage{lipsum}
\usepackage{fancyhdr}       % header
\usepackage{graphicx}       % graphics
\usepackage{subcaption}
\usepackage{mwe}
\usepackage{cancel}

\usepackage{float}
\graphicspath{{media/}}     % organize your images and other figures under media/ folder

\newtheorem{theorem}{Theorem}

\newtheorem{lemma}{Lemma}

\title{A piece-wise constant approximation for non-conjugate Gaussian Process models}

\author{
  Sarem Seitz \\
  Department of Information Systems and Applied Computer Science\\
  Otto-Friedrich-University\\
  Bamberg, Germany\\
  \texttt{sarem.seitz@uni-bamberg.de} \\
}

\begin{document}
\maketitle

\begin{abstract}
Gaussian Processes (GPs) are a versatile and popular method in Bayesian Machine Learning. A common modification are Sparse Variational Gaussian Processes (SVGPs) which are well suited to deal with large datasets. While GPs allow to elegantly deal with Gaussian-distributed target variables in closed form, their applicability can be extended to non-Gaussian data as well. These extensions are usually impossible to treat in closed form and hence require approximate solutions. This paper proposes to approximate the inverse-link function, which is necessary when working with non-Gaussian likelihoods, by a piece-wise constant function. It will be shown that this yields a closed form solution for the corresponding SVGP lower bound. In addition, it is demonstrated how the piece-wise constant function itself can be optimized, resulting in an inverse-link function that can be learnt from the data at hand.
\end{abstract}

% keywords can be removed

\section{Introduction}
\input{sections/00_introduction}

\section{Gaussian Process models and variational approximations}
\input{sections/01_gps}

%\section{Related work}
%\input{sections/02_relatedwork}

\section{Piecewise-constant approximations for non-conjugate likelihood functions}
\input{sections/03_piecewiseapprox}

\section{Experiments}
\input{sections/04_experiments}

\section{Discussion}
\input{sections/05_discussion}

%Bibliography
\bibliographystyle{unsrt}  
\bibliography{references}  

\appendix

\newpage
\section{Proofs}
\input{sections/06_appendix}

\end{document}

%% file: sections/00_introduction.tex
In supervised Bayesian Machine Learning, Gaussian Process (GP) \cite{williams2006gaussian} models are a popular method for dealing with regression and classification tasks. With GPs being stochastic processes, one can utilize a vast body of well-known mathematical properties for practical problems. As GP models are also Bayesian, it possible to encode, in a mathematically sound manner, prior knowledge about a given problem. 

One downside of standard GPs is their cubic scalability in the size of the dataset. This severely limits their applicability in comparison to other popular methods. Popular solutions to this issue typically either optimize the linear operations involved in GP inference or approximate the complex target GP by a simpler one. An instance of the latter are Sparse Variational GPs (SVGPs). The main idea of this approach is the compression of the full dataset into a few so-called \textit{inducing points} ("\textit{sparse}"). A second, \textit{variational}, GP is then conditioned on these inducing points in order to approximate the posterior distribution of the original GP with reduced complexity.

\textbf{Problem.} Another common issue is the non-conjugacy of most non-Gaussian likelihood functions. In most cases, GPs with non-Gaussian likelihoods yield intractable inference that typically require approximation of the underlying quantities. Additionally, non-Gaussian inference demands the choice of an \textit{inverse-link} function to map the GP output to the parameter space of the likelihood function. The choice of inverse-link function can be arbitrary in theory but is typically done so that corresponding inference remains at least numerically tractable. While these two challenges can, for example, be solved in a straightforward manner via Monte-Carlo integration, the resulting estimator could suffer from large variance. Performance of the resulting inference could thus range from inefficient up to almost pointless, depending on the concrete choices of likelihood and inverse-link.

\textbf{Contribution.} To address the issues of the preceding paragraph, this paper proposes to use a piecewise-constant inverse-link function. As can be shown, this results in an analytically tractable lower bound for SVGPs, regardless of the choice of likelihood function. This piecewise inverse-link function can either be fixed to approximate any popular inverse-link function or can be optimized in a data-adaptive manner. Approximation accuracy can be traded-off against computational complexity by increasing or decreasing the amount of piecewise components. Finally, the proposed method is end-to-end differentiable to allow for fast gradient based inference. 

\textbf{Related work.} The treatment of non-conjugate likelihood functions in SVGP models can be divided into two main categories. First, Monte-Carlo samples can be used in order to estimate the log-likelihood expectation in the variational lower bound as first proposed in \cite{DBLP:journals/corr/KingmaW13}. By using the re-parameterization trick as introduced by the latter, it is then possible to optimize the model via gradient-descent. This approach is used for example in \cite{salimbeni2017doubly} in order to tackle an intractable bound in Deep GPs. While MC-estimates are very versatile in regards to the problem of non-conjugate inference, the potentially large variance of the MC estimator can be problematic in practice. \cite{kalos2009monte}.

In regards to approximation of non-conjugate GPs, Laplace approximation methods are arguably the most popular methodology. Contrary to MC-estimation, the Laplace approximation can be used for variational GPs \cite{hensman2015scalable}, as well as non-variational methods \cite{williams2006gaussian}. The application to binary GP classification has been demonstrated for example in \cite{williams2006gaussian}. Other examples include the student-t likelihood \cite{vanhatalo2009gaussian}, or multi-class classification \cite{williams1998bayesian}. 

However, since the Laplace approximation is a second-order Taylor approximation around the posterior mode, there is no straightforward way to improve approximation accuracy by increasing computational resources.

Finally, variational inference \cite{gibbs2000variational, girolami2006variational, khan2012fast} and expectation-maximization (EP) \cite{DBLP:conf/uai/Minka01, williams2006gaussian, jylanki2011robust} are two other classes of algorithms for non-conjugate GP models. These methods are related to each other insofar as they aim to approximate an intractable posterior GP distribution with a surrogate GP. In both cases, a KL divergence between posterior and surrogate distribution is to be minimized. 

\textbf{Outline.} The remainder of this paper is structured as follows: First, we briefly review the basics of GP and SVGP models. Then, we discuss the suggested method and apply it to some extended use-cases. Finally, we conduct experiments to compare the performance of the approach to standard alternatives.

%% file: sections/01_gps.tex
As GP models perform non-parametric Bayesian inference in function space \cite{williams2006gaussian}, we want to find a posterior distribution under a GP prior\footnote{for simplicity, we write $f$ instead of $f(X)$, i.e. we suppress the inputs to $f$}:

\begin{equation}\label{bayeslaw}
    p(f|y)=\frac{p(y|f)p(f)}{\int p(y|f)p(f) df}.
\end{equation}

A GP is uniquely defined by its mean function $m(\cdot):\mathcal{X}\mapsto\mathbb{R}$ and a covariance kernel function $k(\cdot,\cdot):\mathcal{X}\times\mathcal{X}\mapsto \mathbb{R}_0^+$, where $\mathcal{X}$ is a topological space.

We therefore denote the corresponding GP distribution as

\begin{equation}\label{gaussproc}
    p(f)=\mathcal{GP}(f|m(\cdot),k(\cdot,\cdot)).
\end{equation}

While $\mathcal{X}$ can be any topological space, we specify $\mathcal{X}\equiv\mathbb{R}^d$ - hence the corresponding GP is an infinite-dimensional object. As a consequence, there technically exist no density functions $p(f), p(f|y)$. However, since the outlined notation is standard in GP literature, we adapt it for the remainder of this paper. In fact, thinking of $p(f),p(f|y)$ in terms of arbitrary finite dimensional marginal densities of the underlying processes bears no issues for our purposes.

A common choice for the mean function is $m(x)=0\,\,\forall\,\, x\in\mathcal{X}$. For the kernel function we consider the popular Squared Exponential (SE) Kernel,

\begin{equation}\label{ardkern}
    k_{SE}(x,x')=v\cdot exp\left(-0.5 \frac{||x-x'||^2}{s^2}\right),
\end{equation}

with $v > 0$. We denote by $K$ the positive semi-definite \textit{Gram-Matrix}, obtained as $K_{(ij)}=k(x_i,x_j)$, $x_i$ the $i$-th row of the training input matrix $X_N$ containing $N$ observations in total, and write $K_{NN}$ for the Gram-Matrix over $X_N$. $K_{NN}$, together with  $m_{N},\,\, m_{N,(i)}=m(x_i)$, then define covariance matrix and mean vector of the finite dimensional distribution of $f$ at the input locations.

For i.i.d Gaussian observations, we have $p(y|f)=\prod_{i=1}^N \mathcal{N}(y_i|f_i,\sigma^2)$. Thus, we have a Gaussian likelihood whose mean is defined by the marginal distributions of the corresponding GP. The variance parameter is an additional model parameter that needs to be fit to the data. Under this setting, the posterior distribution for a new input matrix $X_*$ can be calculated in closed form:

\begin{equation}\label{postpred}
    p(f_*|y)=\mathcal{N}(f_*|\tilde{\Lambda} y, K_{**}-\tilde{\Lambda} (K_{NN}+I\sigma^2)^{-1} \tilde{\Lambda}^T)
\end{equation}

where $\tilde{\Lambda}=K_{*N}(K_{NN}+I\sigma^2)^{-1}$, $K_{*N,(ij)}=k(x^*_i,x_j)$, $K_{**,(ij)}=k(x^*_i,x^*_j)$; $I$ denotes the identity matrix with according dimension. Again, we circumvent the infinite dimensionality of the posterior by evaluating it at finitely many marginals. As Gaussian-likelihood GPs are fully tractable, the model is usually fitted by maximizing $p(y)=\int p(y|f)p(f) df$ with respect to the kernel and likelihood parameters.

The above amenities of tractable GP inference are quickly shattered when a non-Gaussian likelihood is used. As in the Gaussian case, we presume that one likelihood parameter is modelled by a GP prior. The remaining parameters are treated as fixed. Unless the GP parameter is defined on $\mathbb{R}$, an (element-wise) inverse-link function $g(\cdot)$ is required to map the GP to the corresponding parameter space. This yields the following adjustment to the initial Bayes' law in \eqref{bayeslaw}:

\begin{equation}\label{invlinkpost}
    p(f|y)=\frac{p(y|g(f))p(f)}{\int p(y|g(f))p(f) df}
\end{equation}

Equation \eqref{invlinkpost} is commonly intractable in many settings. We will return to \eqref{invlinkpost} later on and continue with a quick overview on SVGPs.

SVGPs, as demonstrated, particularly, in \cite{titsiassvgp, hensmanbig, hensmansvgpc}, allow  to tackle the issue of intractable posterior inference. At the same time, SVGPs also enable large-scale inference with GP models. We introduce a set of $M$ so called inducing locations $Z_M\subset \mathcal{X}$ and corresponding inducing variables $f_M$. Next, we use a variational GP which is conditioned on these inducing variables, $q(f,f_M)=p(f|f_M)q(f_M)$ - usually $q(f_M)=\mathcal{N}(f_M|a,S),S=L L^T$. The target posterior distribution, $p(f,f_M|y)$, is then approximated through the variational distribution  - by maximizing the \textit{evidence lower bound} (ELBO):

\begin{equation}\label{standardelbo}
    ELBO = \sum_{i=1}^n\mathbb{E}_{p(f|f_M)q(f_M)}\left[\log p(y_{(i)}|f_{(i)})\right]-KL(q(f_M)||p(f_M))
\end{equation}

where we obtain $p(f_M)$ by evaluating the GP prior distribution at inducing locations $Z_M$.
Following standard results for Gaussian random variables, it can also be shown that for marginal $q(f^M_X)$ evaluated at arbitrary $X$ and with $\Lambda=K_{XM}K_{MM}^{-1}$, we have

\begin{equation}\label{marginalvariationalfunction}
    q(f^M_X)=\mathcal{N}(f_|\Lambda a,K_{XX}-\Lambda (K_{MM}-S)^{-1}\Lambda^T)
\end{equation}

From now on, we denote the evaluation of the mean vector in $\eqref{marginalvariationalfunction}$ for given input $X$ as $\mu^{M}_X$ and the corresponding covariance matrix as $\Sigma^M_X$. Single elements are denoted as $\mu^{M}_{X(i)}$ and $\Sigma^M_{X(ij)}$ respectively. Finally, we will write $f^M_{X(i)}$ to denote the $i$-th marginal random variable of $f^M_X$.

%% file: sections/03_piecewiseapprox.tex
We now proceed by using a piece-wise constant function as the inverse-link:

\begin{equation}\label{piececonst}
    g(x)\approx\sum_{k=1}^Kg_k \mathbb{I}_{l_k\leq x< u_k}(x):=\hat{g}_K(x),
\end{equation}

where $g_k\in\mathbb{R}$ and, for all $k<K$, $u_k=l_{k+1}$ and $l_1=-\infty,\,u_K=\infty$. Also, we require, for the corresponding intervals $[l_k,u_k)$, that $\cup_{k=1}^K[l_k,u_k)\equiv\mathbb{R},\,[l_i,u_i)\cap[l_j,u_j)=\emptyset$ for all $i\neq j$. Clearly, $\hat{g}_K(x)$ is a simple function as commonly used in measure theory and can therefore approximate any Borel-measurable function on $\mathbb{R}$ as $K\rightarrow\infty$ \cite{tao2011introduction}. Unfortunately, this is not directly useful for any practical application, where the limit cannot be reached. 

\subsection{Error bounds of piecewise approximations under Gaussian input}

For a measurable function $g:\mathbb{R}\mapsto A\subseteq \mathbb{R}$, we can build its approximation by choosing, for all $k$, $x_k\in[l_k;u_k]$ and setting $g_k=g(x_k)$. Presuming now a Gaussian distribution for a random input variable $X$, we can deduce the following error bound for our approximation (the corresponding proof and all subsequent ones can be found in the appendix):

\begin{theorem}
Let $g:\mathbb{R}\mapsto A\subseteq \mathbb{R}$ be measurable and globally Lipschitz-continuous with Lipschitz-constant $0<\Lambda<\infty$. Also, let $\hat{g}_K$ be as defined in \eqref{piececonst}. Let $p(x)$ denote a Normal distribution with arbitrary mean and variance, $\mu,\,\sigma^2$ and corresponding cumuldative distribution function (c.d.f.), $F(x)$. Also, denote by $\Delta_{p(x)}(b,a)=p(b)-p(a),\,\Delta_{F(x)}(b,a)=F(b)-F(a)$ the difference of the respective functions evaluated at $a,b$. 

We then have

\begin{equation}
\begin{gathered}
    \mathbb{E}_{p(x)}[(g(x)-\hat{g}_K(x))^2] \\
    \leq \Lambda^2 \left\{\mu^2+\sigma^2 - 2\sum_{k=1}^K x_k \left[\mu\Delta_{F(x)}(u_k,l_k)+\sigma\Delta_{p(x)}(u_k,l_k)-\frac{1}{2}x_k\Delta_{F(x)}(u_k,l_k)\right]\right\}.
    \end{gathered}
\end{equation}

and

\begin{equation}
    \lim_{K\rightarrow\infty}\mathbb{E}_{p(x)}[(g(x)-\hat{g}_K(x))^2]=0,
\end{equation}
i.e., the random variable $\hat{g}_K(X)$ converges in mean-square to the random variable $g(X)$ when $X$ is Normal distributed.
\end{theorem}

It might potentially be possible to derive a sharper bound by restricting the functions in question to be bounded and monotonously increasing. This would be advantageous when it comes to popular inverse-link functions like logit and probit. On the other hand, the additional sharpness would be achieved primarily in the tail regions of the underlying distribution. Due to the low density in the tails of the Normal distribution, the actual gain in precision is unlikely to justify further effort.  

As a caveat, Theorem 1 requires the inverse-link function to be globally Lipschitz. This is clearly not fulfilled for the exponential function which is popular for positive parameter spaces. Luckily, it is possible to calculate the mean-squared error with exponential inverse-link in closed form:

\begin{theorem}
    Let $g(x)=exp(x)$ and $\hat{g}_K(x)=\hat{exp}_K(x)=\sum_{k=1}^K\hat{exp}_k\mathbb{I}_{l_k\leq x < u_k}(x)$, where $\hat{exp}_k=exp(x_k)$, with the remaining quantities as in Theorem 1. It then follows that
    
    \begin{equation}
    \begin{gathered}
        \mathbb{E}_{p(x)}\left[(exp(x)-\hat{exp}_K(x))^2\right] \\
        =\sum_{k=1}^K exp\left(2\mu+2\sigma^2\right)\Delta_\Phi\left(-2\sigma+\frac{u_k-\mu}{\sigma},-2\sigma+\frac{l_k-\mu}{\sigma}\right) \\
        -2\cdot\hat{exp}_k\cdot exp\left(\mu+\frac{\sigma^2}{2}\right)\Delta_\Phi\left(-\sigma+\frac{u_k-\mu}{\sigma},-\sigma+\frac{l_k-\mu}{\sigma}\right)  \\
        + \hat{exp}_k^2\Delta_\Phi\left(\frac{u_k-\mu}{\sigma},\frac{l_k-\mu}{\sigma}\right) 
        \end{gathered}
    \end{equation}
    
    where $\Phi(x)$ denotes evaluation of the standard Normal c.d.f. at $x$ with $\Phi(-\infty)=0,\, \Phi(\infty)=1$ and $\Delta_\Phi(b,a)=\Phi(b)-\Phi(a)$.
    
\end{theorem}

Instead of bounding the mean-squared error, we can actually derive it in closed form via sums of  truncated Log-Normal random variables (see proof in the appendix). Clearly, 

$$\lim_{K\rightarrow\infty}\mathbb{E}_{p(x)}\left[(exp(x)-\hat{exp}_K(x))^2\right]=0$$

as $|l_k-u_k|\rightarrow 0$. Depending on which inverse-link is to be used, the above theorems allow to upper bound or calculate exactly the approximation error, depending on the choice of how to set $x_k$. For the remainder of this paper, our choice of $x_k$ is $x_k=u_k$ for all $k<K$ and $x_K=l_K$. Other strategies such as $x_k=(l_k+u_k)/2$ fo all $1<k<K;\, x_1=u_1,\,x_K=l_K$ would be possible but aren't considered further. 

In summary, Theorems 1 and 2 allow to assess the quality of the piece-wise approximation in terms of average squared deviation from the actual value.

\subsection{GP models with piecewise inverse-link functions}

Using a piece-wise constant approximation for the inverse-link function still not solves the problem of an intractable Bayes' formula yet. Rather, we apply the proposed method to SVGPs and show that we obtain an analytically tractable variational lower bound. To see this, consider again the standard ELBO as in \eqref{standardelbo}. We notice that the objective depends on the inverse-link function only via the expectation term. This holds to true regardless on the choice of which inverse-link function is used, as briefly demonstrated in \textbf{Appendix A.4}. The following result then provides the necessary tool for closed-form inference and prediction:

\begin{theorem}
    Let $X_1,...,X_C$ be independent continuous random variables, with probability density functions $p(x_1),...,p(x_C)$ respectively. In addition, let $h(\cdot)$ be a measurable function in the product $\sigma$-algebra corresponding to $X_1,...X_C$. Then, the following holds:

    \begin{equation}\label{mainresult}
        \begin{gathered}
            \mathbb{E}_{p(x_1),...,p(x_C)}\left[h\left(\sum_{k_1=1}^{K_1}g_{k_1}\mathbb{I}_{k_1}(x_1),...,\sum_{k_C=1}^{K_C}g_{k_C}\mathbb{I}_{k_C}(x_C)\right) \right] \\
            =\sum_{k_1,..k_C}^{K_1,...,K_C}h(g_{k_1},...,g_{k_C})P(l_{k_1}<x_1<u_{k_1})\cdots P(l_{k_C}<x_C <u_{k_C})
        \end{gathered}
    \end{equation}

    where $\sum_{i_1,..i_C}^{I_1,...,I_C}$ denotes summation over the cartesian product of all indices.
\end{theorem}

For inference, we can plug \eqref{mainresult} into \eqref{standardelbo} with $h(\cdot)=\log p(y_{(i)}|\cdot)$ and approximating $g(\cdot)\approx\hat{g}_K(\cdot)$ as discussed before. This results in a tractable approximation for the ELBO:

\begin{equation}\label{elboapprox}
    \begin{gathered}
        ELBO = \sum_{i=1}^n\mathbb{E}_{p(f|f_M)q(f_M)}\left[\log p(y_{(i)}|g(f_{(i)}))\right]-KL(q(f_M)||p(f_M)) \\
        \approx \sum_{i=1}^n\sum_{k=1}^{K}\log  p(y_{(i)}|g_k) P(l_k < f^M_{(i)} < u_k) -KL(q(f_M)||p(f_M))
    \end{gathered}
\end{equation}

where $f^M_{(i)}$ denotes the $i$-th marginal random variable of $f$ conditioned on the inducing variables $f_M$. By widening the range for $[u_1,l_K]$ and increasing $K$, the approximation \eqref{elboapprox} can be improved to reliable precision. 

In order to subsequently generate predictions from a trained model for an unseen instance, we derive the posterior predictive distribution at arbitrary input matrix $X$ with $N$ rows for the $i$-th marginal as follows:

\begin{equation}\label{predictivemixture}
    \begin{gathered}
            p(y_{X(i)})=\int\int p(y_{X(i)}|f)p(f|f_M)q(f_M) df df_M = \mathbb{E}_{q(f)}\left[p(y_{X(i)}|f)\right]\\
            =\sum_{k=1}^K  p(y_{X(i)}|g_k) P(l_k < f^M_{X(i)} < u_k)
    \end{gathered}
\end{equation}

This implies that the marginal posterior predictive distributions are mixture distributions with weights determined by the interval probabilities of the marginal variational posterior distributions.  

In theory, we could use the results in this section and \textbf{Theorem 3} in particular to model a likelihood that depends on arbitrarily many parameters modeled by GPs. However, for $T$ independent GPs with $K$ intervals per GP, we would have to calculate $T^K$ probabilities. Clearly, this quickly results in computationally intractable setups. Hence, in the remainder of this paper, we will focus on cases where a single GP - link-function pair is considered. 

Next, we consider three concrete modeling tasks that can be approached with the above results. It should be noted that the proposed method can be applied to any problem involving posterior intractability due to an inverse-link function.

\subsection{Binary GP classification}
\input{sections/031_binary_classification}

\subsection{Heteroscedastic GP regression}
\input{sections/032_heterosced_regression}

\subsection{Learning an inverse-link function}
\input{sections/034_learn_link}

%% file: sections/031_binary_classification.tex
Binary GP classification is a commonly seen problem in regards to non-conjugate GP inference. Here, the GP likelihood term is Bernoulli, i.e.

\begin{equation}\label{bernoullill}
    \begin{gathered}
    p(y|f)=\mathcal{B}er(y|\lambda)=\lambda^y\cdot(1-\lambda)^{(1-y)}\\
    \quad y\in\{0;1\},\lambda\in(0,1)
    \end{gathered}
\end{equation}

This implies an inverse-link function $g:\mathbb{R}\mapsto (0,1)$. A popular choice for $g$ is the sigmoid function

\begin{equation}\label{sigmoidact}
\sigma(f)=\frac{1}{1+e^{-f}}.
\end{equation}

Plugging \eqref{bernoullill} and \eqref{sigmoidact} into \eqref{elboapprox}, we obtain

\begin{equation}\label{elboapprox}
    \begin{gathered}
        \tilde{ELBO}=\sum_{i=1}^N\sum_{k=1}^{K} \left(y_i\log \sigma(x_k) +(1-y_i)\log(1-\sigma(x_k))\right) P(l_k < f^M_{(i)} < u_k) -KL(q(f_M)||p(f_M))
    \end{gathered}
\end{equation}

which can be maximized via gradient ascent over inducing points, inducing variables and kernel hyperparameters. According to \eqref{predictivemixture}, the resulting posterior predictive is then a mixture of Bernoulli distributions.

%% file: sections/032_heterosced_regression.tex
Another application are regression problems with heteroscedastic observation noise. A common use-case of such models are, for example, returns of financial assets whose variance is presumed to be varying over time. Assuming Gaussian observations, we can place two independent GP priors on the mean and variance parameters of the observations. The variance-GP needs to be transformed by an exponential inverse-link function which we can approximate with our method. This results in the following model:

\begin{equation}
    \begin{gathered}
            p(y|f_\mu,f_\sigma) = \mathcal{N}(y|f_\mu,exp(f_\sigma)) \\\\
            p(f_\mu)= \mathcal{GP}(f_\mu|0,k_\mu(x,x')) \\
            p(f_\sigma)=\mathcal{GP}(f_\sigma|0,k_{\sigma}(x,x'))
    \end{gathered}
\end{equation}

The respective posterior distribution,

\begin{equation}
    p(f_\mu,f_\sigma|y)=\frac{p(y|f_\mu,f_\sigma)p(f_\mu)p(f_\sigma)}{p(y)},
\end{equation}

is again intractable. Using SVGPs for both $f_\mu$ and $f_\sigma$, we obtain the lower bound

\begin{equation}
    \begin{gathered}
    ELBO = \sum_{i=1}^n\mathbb{E}_{p(f_\mu|f_{\mu,M})p(f_\sigma|f_{\sigma,M})q(f_{\mu,M})q(f_{\sigma,M})}\left[\log \mathcal{N}(y_i|f_\mu,exp(f_\sigma)) \right]\\
    -KL(q(f_{\mu,M})||p(f_{\mu,M})) -KL(q(f_{\sigma,M})||p(f_{\sigma,M}))
    \end{gathered}
\end{equation}

which, ultimately, can be approximated as

\begin{equation}
    \begin{gathered}
    \hat{ELBO} = \sum_{i=1}^n\sum_{k=1}^{K} \mathbb{E}_{p(f_\mu|f_{\mu,M})q(f_{\mu,M})}\left[\log \mathcal{N}(y_i|f_\mu,exp(x_k)) \right] P(l_k < f_{(i)}^{\sigma,M} < u_k)\\
    -KL(q(f_{\mu,M})||p(f_{\mu,M})) -KL(q(f_{\sigma,M})||p(f_{\sigma,M})) \\
    = \sum_{i=1}^n\sum_{k=1}^{K} \left[\log \mathcal{N}\left(y_i\big|\mu^{\mu,M}_{(i)},exp(x_k)+\Sigma^{\mu,M}_{(ii)}\right)-0.5\frac{\Sigma^{\mu,M}_{(ii)}}{\sigma^2}\right] P(l_k < f_{(i)}^{\sigma,M} < u_k)\\
    -KL(q(f_{\mu,M})||p(f_{\mu,M})) -KL(q(f_{\sigma,M})||p(f_{\sigma,M}))
    \end{gathered}
\end{equation}

%% file: sections/034_learn_link.tex
Typically, when working with non-Gaussian GP likelihoods, a suitable inverse-link function is chosen a-priori. This is obviously a limiting factor when the data does not properly depict the fixed function. In the tradition of modern Machine Learning to 'learn everything from the data', our approach can easily be modified to allow learning the inverse-link approximation as well. It is easily seen, that, most possible likelihood functions are differentiable with respect to the $g_k$. Thus, instead of fixing them based an a-priorily chosen inverse-link function, we treat them as trainable model parameters.

Clearly, this increases the model degrees of freedom considerably for large $K$. To avoid overfitting in this situation, a straightforward solution would be the addition of a regularization term to the ELBO:

\begin{equation}\label{elboreg}
    \tilde{ELBO}_{reg}=\sum_{i=1}^n\sum_{k=1}^{K}\log  p(y_{(i)}|g_k) P(l_k < f^M_{(i)} < u_k) -KL(q(f_M)||p(f_M))-\lambda \sum_{k=1}^{K} (g_k - c)^2
\end{equation}

$c$ could, for example, be chosen as $c=(l_1+u_K)/2$ with $c=0$ if $l_1=-\infty,u_K=\infty$. Another approach could be penalization of the deviation from a pre-defined inverse-link function, i.e.:

\begin{equation}\label{elboreg2}
    \tilde{ELBO}_{reg}=\sum_{i=1}^n\sum_{k=1}^{K}\log  p(y_{(i)}|g_k) P(l_k < f^M_{(i)} < u_k) -KL(q(f_M)||p(f_M))-\lambda \sum_{k=1}^{K} (g_k - g(x_k))^2
\end{equation}

It is obvious that equations \eqref{elboreg} and \eqref{elboreg2} are 'hacky' approaches to regularization and not fully Bayesian. As an actual Bayesian approach, one could additionally place a Gaussian prior over the $g_k$, e.g. $p(g_k)=\mathcal{N}(g_k|\mu_{g_k},\sigma^2_{g_k})$. Approximating the corresponding posterior distributions via $q(g_k)=\mathcal{N}(g_k|\tilde{\mu}_{g_k},\tilde{\sigma}_{g_k}^2)$ and given Gaussian likelihood, one can derive yet another closed-form ELBO:

\begin{equation}\label{elboreg3}
    \begin{gathered}
    \tilde{ELBO}_{reg}=\\
    \sum_{i=1}^n\sum_{k=1}^{K}\left[\log  \mathcal{N}(y_{(i)}|\tilde{\mu}_{g_k},\sigma^2+\tilde{\sigma}_{g_k}) - 0.5\frac{\tilde{\sigma}^2_{g_k}}{\sigma^2}\right] P(l_k < f^M_{(i)} < u_k) -KL(q(f_M)||p(f_M))-\sum_{k=1}^K KL(q(g_k)||p(g_k))
    \end{gathered}
\end{equation}

A proof can be found in the appendix. 

Using this approach allows to model real-valued data with non-Gaussian marginal distributions. Finally, one could also consider $g_k=h_k(x)$ for given input $x\in\mathcal{X}$, i.e. presume that the inverse-link function varies over the input domain. Then, it would be possible to model $h_k(x)$ through a feedforward Neural Network with $K$ output neurons. Respective experiments in the context of this paper did not show any significant performance gains over static $g_k$, so it was not pursued any further.

%% file: sections/04_experiments.tex
To validate the proposed method in a practical context, it was evaluated on standard binary classification and regression datasets.

\textbf{Binary classification.} For binary classification, an approximated sigmoid inverse-link function as in section 3.3 (\textit{SFGP}), GP with learnable piecewise-constant function as in section 3.5 (\textit{SFGP-l}) and the standard GPyTorch implementation of binary SVGP classification are evaluated on both log-likelihood and F1-score. The results thereof are shown in Table 1.\\
\textbf{Regression.}  For regression, a GP with learnable piecewise-constant function as in section 3.5 (\textit{SFGP-l}), a GP with learnable piecewise-constant function and prior function on the approximation parameters as in section 3.5 (\textit{SFGP-Prior}) and the standard GPyTorch implementation of SVGP regression are evaluated on both log-likelihood and RMSE. The results can be found in Table 2.

\textbf{General implementation details.} For all experiments, every continuous variable was standardized by subtracting mean and dividing standard deviation of the training set (\textit{z-standardization}). The respective SVGPs used 10 inducing points and variables, where the locations were drawn from the training data. In the regression case, the GPyTorch benchmark model was initialized with 30 inducing points and variables to balance the additional degrees of freedom for the SFGP models with learnable activation function. For optimization, the \textit{ADAM} optimizer with learning rate $\alpha=0.05$ was used for all models. 

The SFGP models were implemented and evaluated in Julia v1.7 \cite{bezanson2017julia}, using primarily the packages \textit{Flux} \cite{Innes2018} and \textit{Zygote} \cite{DBLP:journals/corr/abs-1907-07587} for automatic differentiation and model training. 

All experiments and the code for the generation of Figures 1 and 2 can be found at \url{https://github.com/SaremS/PieceWiseGP}.

\textbf{Results.} The corresponding tables show the average of the respective evaluation metric over 10-folds of cross-validation and resulting standard deviation. It can be seen that the proposed method generally outperforms the corresponding GPyTorch benchmark on average - in some cases even without any overlap between the intervals of mean $+/-$ one standard deviation (bold table values). In particular, the evaluation log-likelihood is considerably higher in many cases, indicating better distributional fit. RMSE and F1 score, on the other hand, do not differ as much, except for a few dataset. This can be interpreted as the standard GP models already providing reasonable point predictions. 

\begin{table}[H]
\centering
\resizebox{\columnwidth}{!}{%
\begin{tabular}{c|cc|cc|cc}
\multicolumn{1}{c}{}  & \multicolumn{2}{c}{SFGP}  & \multicolumn{2}{c}{SFGP-l} & \multicolumn{2}{c}{GPyTorch}  \\
 & LogLike & F1 & LogLike & F1 & LogLike & F1\\
\hline 
banana & $-0.260\pm 0.020$ & $0.876\pm 0.019$ & $-0.258\pm 0.022$ & $0.876\pm 0.019$ & $-0.272\pm 0.019$ & $0.872\pm 0.014$\\

heart & $-0.397\pm 0.083$ & $0.822\pm 0.075$ & $-0.402\pm 0.091$ & $0.816\pm 0.085$ & $-0.441\pm 0.091$ & $0.819\pm 0.071$\\

diabetes & $\mathbf{-0.241\pm 0.096}$ & $0.912\pm 0.073$ & $\mathbf{-0.233\pm0.091} $ & $0.916\pm0.067 $ & $-0.608\pm0.307$ & $0.703\pm 0.263 $\\

cancer & $-0.093\pm 0.033$ & $0.961\pm 0.031$ & $-0.085\pm 0.033 $ & $0.962\pm0.020 $ & $-0.155\pm0.069$ & $0.954\pm 0.025$\\

\end{tabular}%
}
\caption{Average Log-Likelihood (higher is better) and average F1 score (higher is better) and respective standard-deviations obtained from 10-fold cross validation.}
\label{tab:abc}
\end{table}

\begin{table}[H]
\centering
\resizebox{\columnwidth}{!}{%
\begin{tabular}{c|cc|cc|cc}
\multicolumn{1}{c}{}  & \multicolumn{2}{c}{SFGP-l}  & \multicolumn{2}{c}{SFGP-Prior} & \multicolumn{2}{c}{GPyTorch}  \\
 & LogLike & RMSE & LogLike & RMSE & LogLike & RMSE\\
\hline 
boston & $-0.859\pm 1.099$ & $0.451\pm 0.296$ & $--0.700\pm 0.491$ & $0.471\pm 0.284$ & $-0.853\pm 0.419$ & $0.508\pm 0.269$\\

concrete & $-0.394\pm 0.241$ & $0.348\pm 0.076$ & $-0.473\pm 0.179$ & $0.320\pm 0.096$ & $-0.733\pm 0.109$ & $0.401\pm 0.114$\\

winered & $-1.205\pm 0.081$ & $0.807\pm 0.071$ & $-1.236\pm 0.097$ & $0.842\pm 0.117$ & $-1.226\pm 0.107$ & $0.802\pm 0.064$\\

winewhite & $-1.237\pm 0.083$ & $0.831\pm 0.071$ & $-1.239\pm 0.071$ & $0.839\pm 0.070$ & $-1.277\pm 0.127$ & $0.831\pm 0.074$\\

\end{tabular}%
}
\caption{Average Log-Likelihood (higher is better) and average Root Mean-Squared Error (lower is better) and respective standard-deviations obtained from 10-fold cross validation.}
\label{tab:abc}
\end{table}

\textbf{Optimized inverse-link functions.} For demonstration purposes. Figure 1 shows 'learnt' inverse-link function approximations as explained in section 3.4. For binary classification (Figure 1), the $g_k$ were initialized as $g_k=\sigma(x_k)$ with $\sigma(\cdot)$ as in \eqref{sigmoidact}. It can be seen that the resulting inverse-link functions differ considerably depending on the data used which indicates the usefulness of this approach for better data-fit.

Similarly to classification, Figure 2 shows inverse-link functions for a regression problem with Gaussian likelihood. The left image in Figure 2 depicts an inverse-link function learnt without a regularization term in the ELBO, whereas the right one employed a Bayesian prior regularization term as in \eqref{elboreg3}. The priors for the latter were $p(g_k)=\mathcal{N}(g_k|x_k,1)$ to enforce the inverse-link approximation to be as close to linearity as possible. Clearly, the regularized version (right) is much closer to a linear function (orange line) than the unregularized one (left).

In addition to the posterior mean of the function approximation, the right image in Figure 2 shows the posterior uncertainty intervals (= posterior mean plus 2 posterior standard deviations). The latter increase with increasing distance from zero which implies increasing uncertainty for more extreme values. This can be rationalized by the preceding z-standardization as explained above - most target observations should be centered around zero, hence the model should be most "certain" about values in that region.
\newline

\begin{figure}[H]
        \centering
        \begin{subfigure}[b]{0.4\textwidth}
            \centering
            \includegraphics[width=\textwidth]{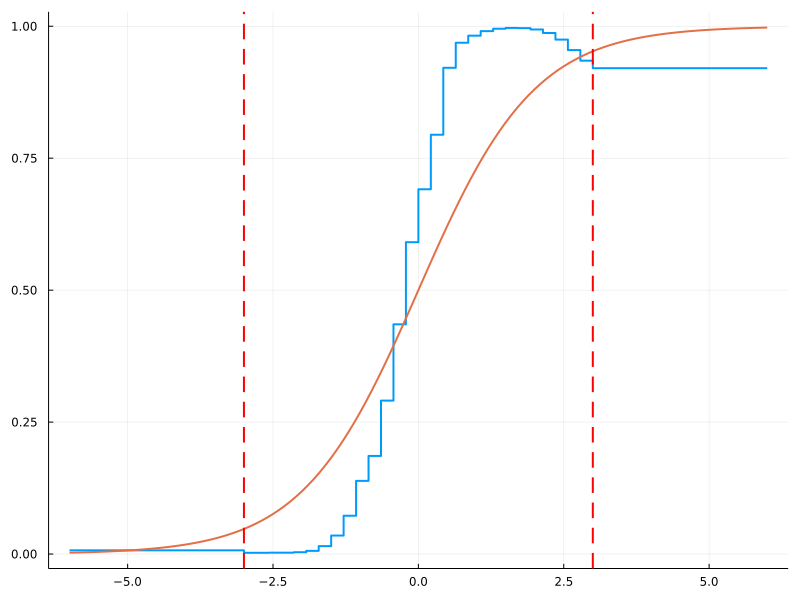}
    
            \label{fig:mean and std of net14}
        \end{subfigure}
        \hfill
        \begin{subfigure}[b]{0.4\textwidth}  
            \centering 
            \includegraphics[width=\textwidth]{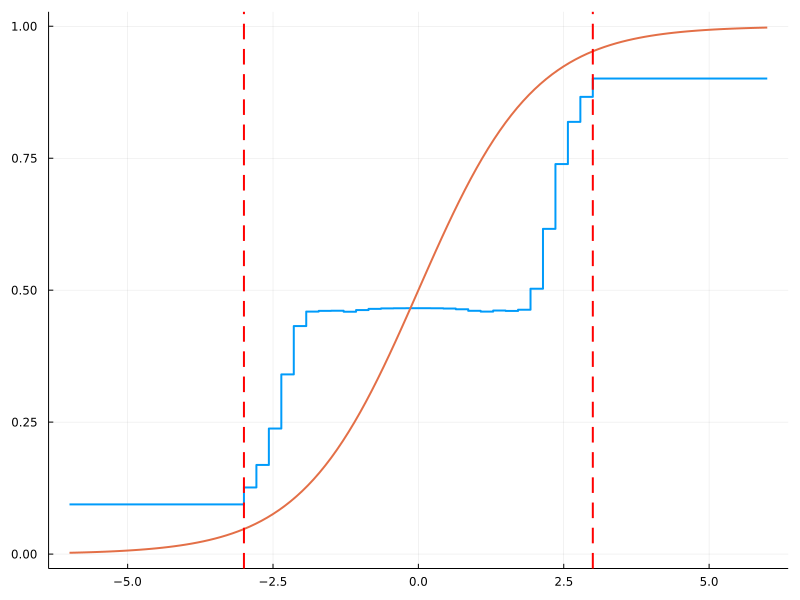}

            \label{fig:mean and std of net24}
        \end{subfigure}
        
        \hfill
        
        \caption[ The average and standard deviation of critical parameters ]
        {{\small Inverse-link function (blue) learnt on \textit{banana} (left) and \textit{heart} (right) with $l_1=-3,u_K=3$ (red) vs. sigmoid inverse-link function (orange)}}  
        \label{fig:invlinkslearnt}
\end{figure}

\begin{figure}[H]
        \centering
        \begin{subfigure}[b]{0.4\textwidth}
            \centering
            \includegraphics[width=\textwidth]{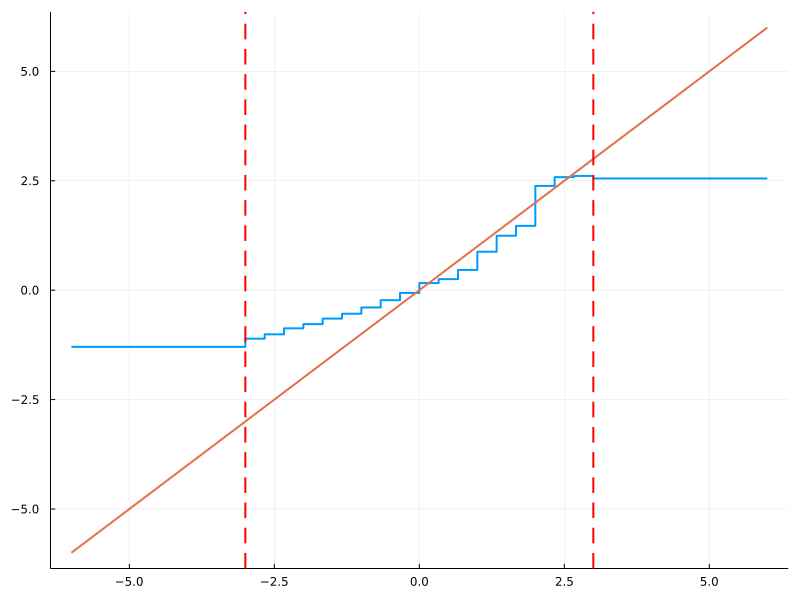}

            \label{fig:mean and std of net14}
        \end{subfigure}
        \hfill
        \begin{subfigure}[b]{0.4\textwidth}  
            \centering 
            \includegraphics[width=\textwidth]{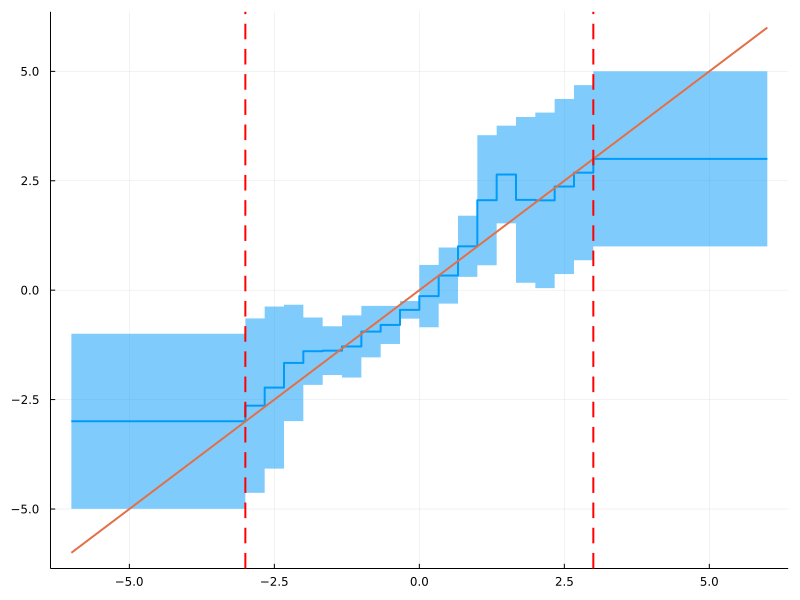}
   
            \label{fig:mean and std of net24}
        \end{subfigure}
        \vskip\baselineskip

        \caption[ The average and standard deviation of critical parameters ]
        {\small Inverse-link function (blue) learnt on \textit{banana} (left; no regularization) and \textit{heart} (right, Bayesian regularization) with $l_1=-3,u_K=3$ (red) vs. sigmoid inverse-link function (orange)} 
        \label{fig:mean and std of nets}
\end{figure}

%% file: sections/05_discussion.tex
This work presented a new method for working with non-Gaussian GP likelihoods by approximating the necessary inverse-link function by a piecewise-constant one. As a result, all inference becomes analytically tractable when used in conjunction with a SVGP model. In addition it is possible to learn an approximation function from the data itself, removing the need to select a suitable inverse-link function a-priori. Regarding benchmarks, the proposed method achieves reasonable performance gains over existing implementations.

The primary limitation, on the other hand, is the combinatorical limitation when extending the method to multiple GPs in the same model. This makes the method, in its current form, not useful for n-ary classification which is, however, an important problem in modern Machine Learning. Future work in this direction could therefore help to make the proposed method more applicable.

%% file: sections/06_appendix.tex
The primary results rely on the following Lemma:

\begin{lemma}
    Let $p(x)$ denote the probability density of a continuous random variable with c.d.f. $F(x)$, it then follows that
    $$\int_a^b x p(x)dx = \Delta_{F(x)}(b,a) \mathbb{E}_{p(x|a\leq x < b)}[X]$$
    where $p(x|a\leq x \leq b)$ denotes the probability density of the $X$ truncated within $(a,b)$\newline
    
    In particular
    
    $$\int_a^b x \mathcal{N}(x|\mu,\sigma^2)dx = \Delta_{F(x)}(b,a)\left(\mu + \sigma\frac{\Delta_{p(x)}(b,a)}{\Delta_{F(x)}(b,a)}\right)= \mu\Delta_{F(x)}(b,a) + \sigma\Delta_{p(x)}(b,a)$$\newline
\end{lemma}

\textit{Proof}    This follows directly via

$$\int_a^b x p(x)dx=\Delta_{F(x)}(b,a) \int_a^b \frac{xp(x)}{\Delta_{F(x)}(b,a)}dx=\Delta_{F(x)}(b,a) \mathbb{E}_{p(x|a\leq x < b)}[X]$$

\subsection{Theorem 1}

In order to prove this result, we apply the assumptions made and then simplify:

$$\mathbb{E}[(g(x)-\hat{g}_K(x))^2]$$

$$=\int_{-\infty}^\infty \left(g(x)-\sum_{k=1}^K g_k\mathbb{I}_{l_k\leq x < u_k}(x)\right)^2 p(x)dx$$

$$=\sum_{k=1}^K\int_{l_k}^{u_k} \left(g(x)-g_k\right)^2 p(x)dx$$

$$\text{(splitting up the integral leaves only the corresponding indicator function take non-zero value)}$$

$$\leq\Lambda^2\sum_{k=1}^{K}\int_{l_k}^{u_k} \left(x-x_k\right)^2 p(x)dx $$

$$\text{(via Lipschitz-continuity)}$$

$$=\Lambda^2\left\{\sum_{k=1}^{K}\int_{l_k}^{u_k} \left[x^2 - 2x_kx + x_k^2 \right]p(x)dx \right\}$$

\begin{equation}\label{refstep}
    =\Lambda^2 \left\{\mathbb{E}\left[X^2\right]-2\sum_{k=1}^K x_k\int_{l_k}^{u_k}xp(x)dx+\sum_{k=1}^K x_k^2 \int_{l_k}^{u_k}p(x)dx\right\}
\end{equation}

$$=\Lambda^2 \left\{\mu^2+\sigma^2 - 2\sum_{k=1}^K x_k \left[\mu\Delta_{F(x)}(u_k,l_k)+\sigma\Delta_{p(x)}(u_k,l_k)-\frac{1}{2}x_k\Delta_{F(x)}(u_k,l_k)\right]\right\}$$

$$\text{(by applying Lemma 1)}$$

To prove mean-square convergence, we set, for fixed $k$, $l_k = x_k-\frac{i_{l_k}}{K},\,u_k = x_k+\frac{i_{u_k}}{K}$ 

$$\lim_{K\rightarrow\infty}\sum_{k=1}^K x_k \int_{l_k}^{u_k}xp(x)dx=\lim_{K\rightarrow\infty}\sum_{k=1}^K \int_{\Big[x_k-\frac{i_{l_k}}{K},x_k+\frac{i_{u_k}}{K}\Big)}x_k\cdot x p(x)dx$$

$$=\sum_{k=1}^\infty x_k^2p(x_k)=\int x^2p(x)dx=\mathbb{E}\left[X^2\right]$$

By the same argument, we obtain

$$\lim_{K\rightarrow\infty}\sum_{k=1}^K x_k^2 \int_{l_k}^{u_k}p(x)dx=\mathbb{E}\left[X^2\right]$$

and therefore

$$\lim_{K\rightarrow\infty}\sum_{k=1}^K\Lambda^2 \left\{\mathbb{E}\left[X^2\right]-2\sum_{k=1}^K x_k\int_{l_k}^{u_k}xp(x)dx+\sum_{k=1}^K x_k^2 \int_{l_k}^{u_k}p(x)dx\right\}$$

$$= \Lambda^2 \left\{\mathbb{E}\left[X^2\right]-2\mathbb{E}\left[X^2\right]+\mathbb{E}\left[X^2\right]\right\}=0$$
$\square$

\subsection{Theorem 2}

We now explicitly denote the densities of a Normal and corresponding Log-Normal distribution with parameters $\mu,\sigma^2$ as $\mathcal{N}(x|\mu,\sigma^2)$ and $\mathcal{L}og\mathcal{N}(x|\mu,\sigma)$. The principal idea behind the proof is to re-arrange the terms until we arrive at means of truncated log-normal random variables. For the latter, there exists a closed form solution:

$$\mathbb{E}\left[(exp(x)-\hat{exp}_K(x))^2\right]$$

$$=\int_{-\infty}^\infty\left(exp(x)-\hat{exp}_K(x)\right)^2 \mathcal{N}(x|\mu,\sigma^2) dx$$

$$=\sum_{k=1}^K\int_{l_k}^{u_k}\left(exp(x)-\hat{exp}_k\right)^2 \mathcal{N}(x|\mu,\sigma^2) dx$$

$$\text{(splitting up the integral as in Theorem 1)}$$

$$=\sum_{k=1}^K\int_{l_k}^{u_k}exp(2x)\mathcal{N}(x|\mu,\sigma^2) dx-2\cdot\hat{exp}_k\int_{l_k}^{u_k}exp(x)\mathcal{N}(x|\mu,\sigma^2) dx+ \hat{exp}_k^2\int_{l_k}^{u_k}\mathcal{N}(x|\mu,\sigma^2) dx$$

$$=\sum_{k=1}^K\int_{exp(2l_k)}^{exp(2u_k)}y\cdot\mathcal{L}og\mathcal{N}(y|2\mu,2\sigma) dy-2\cdot\hat{exp}_k\int_{exp(l_k)}^{exp(u_k)}z\cdot\mathcal{L}og\mathcal{N}(z|\mu,\sigma) dz+ \hat{exp}_k^2\int_{l_k}^{u_k}\mathcal{N}(x|\mu,\sigma^2) dx$$

$$\text{(via change of variables)}$$

\begin{gather*}
=\sum_{k=1}^K P(l_k\leq x \leq u_k)\cdot\Bigg[\int_{exp(2l_k)}^{exp(2u_k)}\frac{y\cdot\mathcal{L}og\mathcal{N}(y|2\mu,2\sigma)}{P(exp(2 l_k)\leq y \leq exp(2 u_k))} dy \\
-2\cdot\hat{exp}_k\int_{exp(l_k)}^{exp(u_k)}\frac{z\cdot\mathcal{L}og\mathcal{N}(z|\mu,\sigma)}{P(exp( l_k)\leq z \leq exp(u_k))} dz+ \hat{exp}_k^2\Bigg]
\end{gather*}

$$\text{(re-transforming the denominators cancels out the left-hand factor)}$$

\begin{gather*}
=\sum_{k=1}^K P(l_k\leq x \leq u_k)\cdot\Bigg[exp\left(2\mu+\frac{4\sigma^2}{2}\right)\frac{\Phi\left(-2\sigma+\frac{log(exp(2u_k))-2\mu}{2\sigma}\right)-\Phi\left(-2\sigma+\frac{log(exp(2l_k))-2\mu}{2\sigma}\right)}{\Phi\left(\frac{log(exp(2u_k))-2\mu}{2\sigma}\right)-\Phi\left(\frac{log(exp(2l_k))-2\mu}{2\sigma}\right)} \\
-2\cdot\hat{exp}_k\cdot exp\left(\mu+\frac{\sigma^2}{2}\right)\frac{\Phi\left(-\sigma+\frac{log(exp(u_k))-\mu}{\sigma}\right)-\Phi\left(-\sigma+\frac{log(exp(l_k))-\mu}{\sigma}\right)}{\Phi\left(\frac{log(exp(u_k))-\mu}{\sigma}\right)-\Phi\left(\frac{log(exp(l_k))-\mu}{\sigma}\right)}  \\
+ \hat{exp}_k^2\Bigg]
\end{gather*}

\begin{gather*}\text{(the first two summands are means over truncated log-normal random variables}\\ \text{whose mean is available in closed form - see e.g. \cite{wang2012speed} for a formal proof; also notice the explicit use of Lemma 1)}
\end{gather*}

\begin{gather*}
=\sum_{k=1}^K \Delta_\Phi\left(\frac{u_k-\mu}{\sigma},\frac{l_k-\mu}{\sigma}\right)\cdot\Bigg[exp\left(2\mu+2\sigma^2\right)\frac{\Delta_\Phi\left(-2\sigma+\frac{u_k-\mu}{\sigma},-2\sigma+\frac{l_k-\mu}{\sigma}\right)}{\Delta_\Phi\left(\frac{u_k-\mu}{\sigma},\frac{l_k-\mu}{\sigma}\right)} \\
-2\cdot\hat{exp}_k\cdot exp\left(\mu+\frac{\sigma^2}{2}\right)\frac{\Delta_\Phi\left(-\sigma+\frac{u_k-\mu}{\sigma},-\sigma+\frac{l_k-\mu}{\sigma}\right)}{\Delta_\Phi\left(\frac{u_k-\mu}{\sigma},\frac{l_k-\mu}{\sigma}\right)}  \\
+ \hat{exp}_k^2\Bigg]
\end{gather*}

\begin{gather*}
=\sum_{k=1}^K exp\left(2\mu+2\sigma^2\right)\Delta_\Phi\left(-2\sigma+\frac{u_k-\mu}{\sigma},-2\sigma+\frac{l_k-\mu}{\sigma}\right) \\
-2\cdot\hat{exp}_k\cdot exp\left(\mu+\frac{\sigma^2}{2}\right)\Delta_\Phi\left(-\sigma+\frac{u_k-\mu}{\sigma},-\sigma+\frac{l_k-\mu}{\sigma}\right)  \\
+ \hat{exp}_k^2\Delta_\Phi\left(\frac{u_k-\mu}{\sigma},\frac{l_k-\mu}{\sigma}\right) 
\end{gather*}
$\square$

\subsection{Proposition 3}

This result is derived by, again, splitting up integrals alongside the respective indicator functions.

$$\int\cdots\int f\left(\sum_{k_1=1}^{K_1}g_{k_1}\mathbb{I}_{k_1}(x_1),...,\sum_{k_C=1}^{K_C}g_{k_C}\mathbb{I}_{k_C}(x_C)\right)p(x_1)\cdots p(x_C)\cdot dx_1\cdots dx_C$$

$$=\int\cdots \sum_{k_1=1}^{K_1} \int_{l_{k_1}}^{u_{k_1}} f\left(g_{k_1}\cdot 1 +0,...,\sum_{k_C=1}^{K_C}g_{k_C}\mathbb{I}_{k_C}(x_C)\right)p(x_1)\cdots p(x_C)\cdot dx_1\cdots dx_C$$

$$=\int\cdots \int \sum_{k_1=1}^{K_1} f\left(g_{k_1},\sum_{k_2=1}^{K_2}g_{k_2}\mathbb{I}_{k_2}(x_2),...,\sum_{k_C=1}^{K_C}g_{k_C}\mathbb{I}_{k_C}(x_C)\right)\int_{l_{k_1}}^{u_{k_1}} p(x_1) dx_1\cdot p(x_2)\cdots p(x_C)\cdot dx_2\cdots dx_C $$

$$=\int\cdots \int \sum_{k_1=1}^{K_1} f\left(g_{k_1},\sum_{k_2=1}^{K_2}g_{k_2}\mathbb{I}_{k_2}(x_2),...,\sum_{k_C=1}^{K_C}g_{k_C}\mathbb{I}_{k_C}(x_C)\right)P(l_{k_1}\leq x_1 \leq u_{k_1})\cdot p(x_2)\cdots p(x_C)\cdot dx_2\cdots dx_C $$

$$\vdots$$

$$=\sum_{k_1,..k_C}^{K_1,...,K_C}f(g_{k_1},...,g_{k_C})P(l_{k_1}\leq,x_1,u_{k_1})\cdots P(l_{k_C}\leq,x_C,u_{k_C})$$
$\square$

\subsection{$\hat{ELBO}_{reg}$ in equation \eqref{elboreg3}}

We derive a GP-ELBO as usual but include a second, arbitrary set of additional random variables that is summarized as $A$:

$$0\leq KL(q(f,f_M,A)||p(f,f_M,A|y))$$

$$=KL(p(f|f_M)q(f_M)q(A)||p(f,A|y))$$

$$=\mathbb{E}_{p(f|f_M)q(f_M)q(A)} \left[\log\frac{p(f|f_M)q(f_M)q(A)}{p(f,f_M,A|y)}\right]$$

$$=\mathbb{E}_{p(f|f_M)q(f_M)q(A)} \left[\log\frac{p(f|M)q(f_M)q(A)\cdot p(y)}{p(y|f,f_M,A)\cdot p(f|f_M)p(f_M)p(A)}\right]$$

$$=\mathbb{E}_{p(f|f_M)q(f_M)q(A)} \left[\log\frac{\cancel{p(f|f_M)}q(f_M)q(A)\cdot p(y)}{p(y|f,f_M,A)\cdot \cancel{p(f|f_M)}p(f_M)p(A)}\right]$$

$$=KL(q(f_M)||p(f_M)) + KL(q(f_M)||p(f_M)) - \mathbb{E}_{p(f|f_M)q(f_M)q(A)}\left[\log p(y|f,f_M,A)\right] + \log p(y)$$

$$\Rightarrow  \log p(y)\geq \mathbb{E}_{p(f|f_M)q(f_M)q(A)}\left[\log p(y|f,A)\right] - KL(q(f_M)||p(f_M)) - KL(q(A)||p(A))$$

The latter depends on the inverse-link function only in the expecation on the left. Also, notice that $\mathbb{E}_{p(f|f_M)q(f_M)}\left[\log p(y|f,f_M,A)\right]=\mathbb{E}_{q(f)}\left[\log p(y|f,A)\right]$ as $y$ is independent of $f_M$ when conditioned on $f$.

Plugging in the $g_1,...,g_K$ as $A$ with prior distribution $p(A)\equiv p(g_1,...,g_K)=p(g_1)\cdots p(g_K)=\prod_{k=1}^K\mathcal{N}(g_k|\mu_{g_k},\sigma^2_{g_k})$ and variational distribution $q(A)\equiv q(g_1,...,g_K)=q(g_1)\cdots q(g_K)=\prod_{k=1}^K\mathcal{N}(g_k|\tilde{\mu}_{g_k},\tilde{\sigma}^2_{g_k})$ then yield an analytically tractable ELBO:

$$\hat{ELBO}_{reg}=\mathbb{E}_{p(f|f_M)q(f_M)q(A)}\left[\log p(y|f,A)\right] - KL(q(f_M)||p(f_M)) - KL(q(A)||p(A))$$

$$\equiv\mathbb{E}_{p(f|f_M)q(f_M)q(g_1)\cdots q(g_K)}\left[\log \mathcal{N}\left(y\bigg|\sum_{k=1}^K g_k \mathbb{I}_{l_k\leq x < u_k}(f),\sigma^2\right)\right] - KL(q(f_M)||p(f_M)) - KL(q(g_1)\cdots q(g_K)||p(g_1)\cdot p(g_K))$$

$$=\sum_{i=1}^n\mathbb{E}_{q(f^M)q(g_1)\cdots q(g_K)}\left[\log \mathcal{N}\left(y_i\bigg|\sum_{k=1}^K g_k \mathbb{I}_{l_k\leq x < u_k}\left(f^M_{(i)}\right),\sigma^2\right)\right] -KL(q(f_M)||p(f_M))-\sum_{k=1}^K KL(q(g_k)||p(g_k))$$

$$=\sum_{i=1}^n\mathbb{E}_{q(g_1)\cdots q(g_K)}\left[\sum_{k=1}^{K}\log \mathcal{N}\left(y_i|g_k ,\sigma^2\right)P(l_k \leq f^M_{(i)} < u_k)\right] -KL(q(f_M)||p(f_M))-\sum_{k=1}^K KL(q(g_k)||p(g_k))$$

$$=\sum_{i=1}^n\sum_{k=1}^{K}\left[\log \mathcal{N}\left(y_i|\tilde{\mu}_{g_k} ,\sigma^2+\tilde{\sigma}^2_{g_k}\right)-0.5\frac{\tilde{\sigma}^2_{g_k}}{\sigma^2}\right]P(l_k \leq f^M_{(i)} < u_k) -KL(q(f_M)||p(f_M))-\sum_{k=1}^K KL(q(g_k)||p(g_k))$$ 

$\square$